\pdfoutput=1

\documentclass[11pt,letterpaper]{article}
\usepackage{emnlp2016}
\usepackage{times}
\usepackage{latexsym}

\usepackage{amsmath}
\usepackage{lipsum}
\usepackage{makecell}
\usepackage{times}
\usepackage{latexsym}
\usepackage[latin1]{inputenc}
\usepackage{amsfonts}
\usepackage{amssymb}
\usepackage{graphicx}
\usepackage[noend]{algpseudocode}
\usepackage{listings}
\usepackage{color}
\usepackage{booktabs}
\usepackage{array}
\usepackage{tabularx}
\usepackage{setspace}
\usepackage{hyperref}
\usepackage{float}
\usepackage{multirow}
\usepackage[labelfont=bf]{caption}
\usepackage[T1]{fontenc}
\usepackage{inconsolata}
\usepackage{listings}
\usepackage{color}
\usepackage{tikz}
\restylefloat{table}
\usepackage{enumitem}
\usepackage{xcolor}
\usepackage{subfigure}

\emnlpfinalcopy

\def\emnlppaperid{***}

\title{Gated Word-Character Recurrent Language Model}

\author{Yasumasa Miyamoto\\Center for Data Science\\New York University\\
  {\tt yasumasa.miyamoto@nyu.edu}
\And
  Kyunghyun Cho\\ Courant Institute of \\ Mathematical Sciences\\ \& Centre for Data Science \\New York University \\ 
  {\tt kyunghyun.cho@nyu.edu}}

\date{}

\begin{document}

\maketitle

\begin{abstract}
We introduce a recurrent neural network language model (RNN-LM) with long short-term memory (LSTM) units that utilizes both character-level and word-level inputs. Our model has a gate that adaptively finds the optimal mixture of the  character-level and word-level inputs.  The gate creates the final vector representation of a word by combining two distinct representations of the word. The character-level inputs are converted into vector representations of words using a bidirectional LSTM. The word-level inputs are projected into another high-dimensional space by a word lookup table. 
The final vector representations of words are used in the LSTM language model which predicts the next word given all the preceding words. Our model with the gating mechanism effectively utilizes the character-level inputs for rare and out-of-vocabulary words and outperforms word-level language models on several English corpora. 
\end{abstract}

\section{Introduction}
Recurrent neural networks (RNNs) achieve state-of-the-art performance on fundamental tasks of natural language processing (NLP) such as language modeling (RNN-LM) \cite{DBLP:journals/corr/JozefowiczVSSW16,zoph2016simple}. RNN-LMs are usually based on the word-level information or subword-level information such as characters \cite{mikolov2012subword}, and predictions are made at either word level or subword level respectively. 

In word-level LMs, the probability distribution over the vocabulary conditioned on preceding words is computed at the output layer using a softmax function. \footnote{softmax function is defined as $f(x_i) = \frac{\exp{x_i}}{\sum_{k}^{} \exp{x_k}}. $} Word-level LMs require a predefined vocabulary size since the computational complexity of a softmax function grows with respect to the vocabulary size. This closed vocabulary approach tends to ignore rare words and typos, as the words do not appear in the vocabulary are replaced with an out-of-vocabulary (OOV) token. The words appearing in vocabulary are indexed and associated with high-dimensional vectors. This process is done through a word lookup table.

Although this approach brings a high degree of freedom in learning expressions of words, information about morphemes such as prefix, root, and suffix is lost when the word is converted into an index. Also, word-level language models require some heuristics to differentiate between the OOV words, otherwise it assigns the exactly same vector to all the OOV words. These are the major limitations of word-level LMs. 

In order to alleviate these issues, we introduce an RNN-LM that utilizes both character-level and word-level inputs. In particular, our model has a gate that adaptively choose between two distinct ways to represent each word: a word vector derived from the character-level information and a word vector stored in the word lookup table.  This gate is trained to make this decision based on the input word. 

According to the experiments, our model with the gate outperforms other models on the Penn Treebank (PTB), BBC, and IMDB Movie Review datasets. Also, the trained gating values show that the gating mechanism effectively utilizes the character-level information when it encounters rare words. 

\vspace{4pt}

\noindent \textbf{Related Work}$\:\:$  Character-level language models that make word-level prediction have recently been proposed.  Ling et al. \shortcite{wang:2015} introduce the compositional character-to-word (C2W) model that takes as input character-level representation of a word and generates vector representation of the word using a bidirectional LSTM \cite{DBLP:journals/nn/GravesS05}. Kim et al. \shortcite{DBLP:journals/corr/KimJSR15} propose a convolutional neural network (CNN) based character-level language model and achieve the state-of-the-art perplexity on the PTB dataset with a significantly fewer parameters. 
 
Moreover, word--character hybrid models have been studied on different NLP tasks. Kang et al. \shortcite{DBLP:conf/interspeech/KangNN11} apply a word--character hybrid language model on Chinese using a neural network language model \cite{DBLP:journals/jmlr/BengioDVJ03}. Santos and Zadrozny \shortcite{DBLP:conf/icml/SantosZ14} produce high performance part-of-speech taggers using a deep neural network that learns character-level representation of words and associates them with usual word representations. Bojanowski et al. \shortcite{DBLP:journals/corr/BojanowskiJM15} investigate RNN models that predict characters based on the character and word level inputs. Luong and Manning \shortcite{DBLP:journals/corr/LuongM16} present word--character hybrid neural machine translation systems that consult the character-level information for rare words.

\section{Model Description}

\begin{figure}[t]
	\centering
	\includegraphics[width=1.0\linewidth]{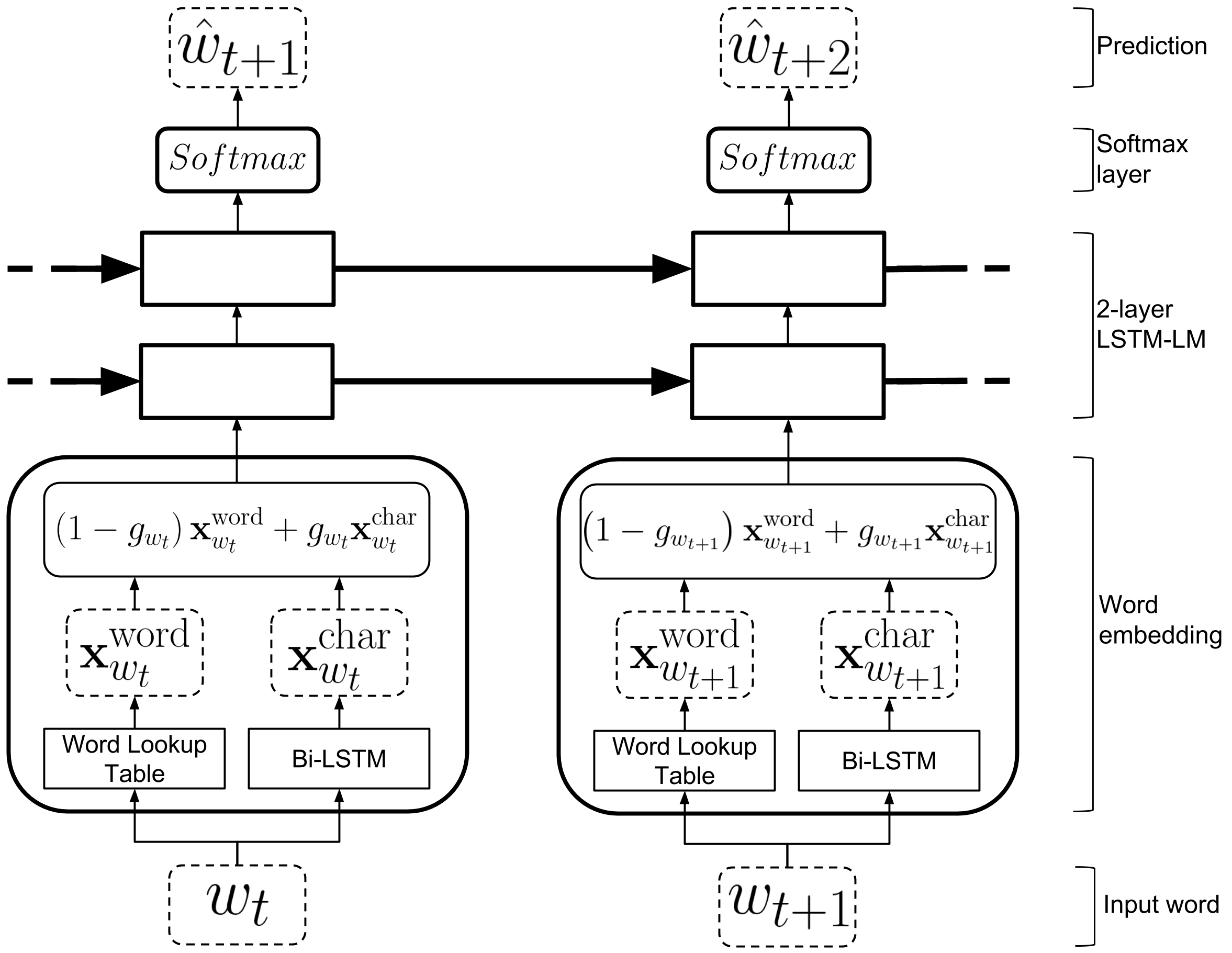}
	\caption{The model architecture of the gated word-character recurrent language model. $w_t$ is an input word at $t$. $\mathbf{x}^{\text{word}}_{w_t}$ is a word vector stored in the word lookup table. $\mathbf{x}^{\text{char}}_{w_t}$ is a word vector derived from the character-level input.  $g_{w_t}$ is a gating value of a word $w_t$.  $\hat{w}_{t+1}$ is a prediction made at $t$.}
	\label{fig:lm}
\end{figure}

The model architecture of the proposed word--character hybrid language model is shown in Fig.~\ref{fig:lm}. 
\vspace{4pt}

\noindent \textbf{Word Embedding}$\:\:$  At each time step $t$, both the word lookup table  and a bidirectional LSTM take the same word $w_t$ as an input.  The word-level input is projected into a high-dimensional space by a word lookup table $\mathbf{E} \in \mathbb{R}^{|V| \times d}$, where $|V|$ is the vocabulary size and $d$ is the dimension of a word vector: 
\begin{align}
\mathbf{x}_{w_t}^{\text{word}} = \mathbf{E}^{\top} \mathbf{w}_{w_t},
\end{align}
where $\mathbf{w}_{w_t} \in \mathbb{R}^{|V|}$ is a one-hot vector whose $i$-th element is $1$, and other elements are $0$.
The character--level input is converted into a word vector by using a bidirectional LSTM. The last hidden states of forward and reverse recurrent networks are linearly combined:
\begin{align}
\mathbf{x}_{w_t}^{\text{char}} = \mathbf{W}^f \mathbf{h}^f_{w_t}+ \mathbf{W}^r \mathbf{h}^r_{w_t} + \mathbf{b},
\end{align}
where $\mathbf{h}^f_{w_t}, \mathbf{h}^r_{w_t} \in \mathbb{R}^{d}$ are the last states of the forward and the reverse  LSTM respectively. $ \mathbf{W}^f, \mathbf{W}^r  \in \mathbb{R}^{d \times d}$ and  $\mathbf{b} \in \mathbb{R}^{d}$ are trainable parameters, and  $\mathbf{x}_{w_t}^{\text{char}} \in \mathbb{R}^{d}$ is the vector representation of the word $w_t$ using a character input. The generated vectors $\mathbf{x}_{w_t}^{\text{word}}$ and $\mathbf{x}_{w_t}^{\text{char}}$  are mixed by a gate $g_{w_t}$ as 
\begin{equation} \label{gate}
\begin{aligned}
g_{w_t} &= \sigma\left( \mathbf{v}^{\top}_g \mathbf{x}^{\text{word}}_{w_t}  +  b_g\right)\\
\mathbf{x}_{w_t} &= \left( 1 -  g_{w_t} \right)  \mathbf{x}^{\text{word}}_{w_t} + g_{w_t}  \mathbf{x}^{\text{char}}_{w_t},
\end{aligned}
\end{equation}

where $\mathbf{v}_g  \in \mathbb{R}^{d}$ is a weight vector, $ b_g \in \mathbb{R}$ is a bias scalar, $\sigma (\cdot)$ is a sigmoid function. This gate value is independent of a time step. Even if a word appears in different contexts, the same gate value is applied. Hashimoto and Tsuruoka \shortcite{DBLP:journals/corr/HashimotoT16} apply a very similar approach to compositional and non-compositional phrase embeddings and achieve state-of-the-art results on compositionality detection and verb disambiguation tasks.

\renewcommand{\arraystretch}{1}
\begin{table*}[t]
	\centering
	\small
	\begin{tabular}{l||c c | c c| c c}
		\hline
		\textbf{}& \multicolumn{2}{c|}{\textbf{PTB}}& \multicolumn{2}{c|}{\textbf{BBC}}& \multicolumn{2}{c}{\textbf{IMDB}}\\
		\cline{2-7}
		\multicolumn{1}{c||}{\textbf{Model}}& \multicolumn{1}{c}{\textbf{Validation}}& \multicolumn{1}{c|}{\textbf{Test}}& \multicolumn{1}{c}{\textbf{Validation}}& \multicolumn{1}{c|}{\textbf{Test}}& \multicolumn{1}{c}{\textbf{Validation}}& \multicolumn{1}{c}{\textbf{Test}} \\
		\hline
		Gated Word \& Char, adaptive &117.49&113.87&\textbf{78.56}&\textbf{87.16}&71.99&72.29\\
		Gated Word \& Char, adaptive (Pre-train)&117.03&112.90&80.37&87.51&71.16&71.49\\
		\hline
		Gated Word \& Char, $g = 0.25$  & 119.45& 115.55 & 79.67 & 88.04 & 71.81 & 72.14\\
		Gated Word \& Char, $g = 0.25$  (Pre-train) & \textbf{117.01} & \textbf{113.52} & 80.07 & 87.99 & \textbf{70.60} & \textbf{70.87}\\
		Gated Word \& Char, $g = 0.5$  & 126.01 & 121.99 & 89.27 & 94.91 & 106.78 & 107.33\\
		Gated Word \& Char, $g = 0.5$  (Pre-train) & 117.54 & 113.03 & 82.09 & 88.61 & 109.69 & 110.28\\
		Gated Word \& Char, $g = 0.75$  & 135.58 & 135.00 & 105.54 & 111.47 & 115.58 & 116.02\\
		Gated Word \& Char, $g = 0.75$  (Pre-train) & 179.69 & 172.85 & 132.96 & 136.01 & 106.31 & 106.86\\
		\hline
		Word Only&118.03&115.65&84.47&90.90&72.42&72.75\\
		Character Only&132.45&126.80&88.03&97.71&98.10&98.59\\
		Word \& Character&125.05&121.09&88.77&95.44&77.94&78.29\\
		Word \& Character  (Pre-train)&122.31&118.85&84.27&91.24&80.60&81.01\\
		\hline
		Non-regularized LSTM (Zaremba, 2014)& 120.7 & 114.5 & - & - & - & -\\
		\hline
	\end{tabular}
	\caption{Validation and test perplexities on Penn Treebank (PTB), BBC, IMDB Movie Reviews datasets. }\label{tab:perp}
\end{table*}

\vspace{4pt}	
\noindent \textbf{Language Modeling}$\:\:$  The output vector $\mathbf{x}_{w_t}$ is used as an input to a LSTM language model. Since the word embedding part is independent from the language modeling part, our model retains the flexibility to change the architecture of the language modeling part.  We use the architecture similar to the non-regularized LSTM model by Zaremba et al. \shortcite{DBLP:journals/corr/ZarembaSV14}. One step of LSTM computation corresponds to     
\begin{equation} \label{lstm}
\begin{aligned}
\mathbf{f}_t &= \sigma\left( \mathbf{W}_f \mathbf{x}_{w_t} + \mathbf{U}_f \mathbf{h}_{t-1}  + \mathbf{b}_f \right)\\
\mathbf{i}_t &= \sigma\left( \mathbf{W}_i \mathbf{x}_{w_t} + \mathbf{U}_i \mathbf{h}_{t-1}  + \mathbf{b}_i  \right)\\
\mathbf{\tilde{c}}_t &= \tanh\left( \mathbf{W}_{\tilde{c}} \mathbf{x}_{w_t} + \mathbf{U}_{\tilde{c}} \mathbf{h}_{t-1}  + \mathbf{b}_{\tilde{c}} \right)\\
\mathbf{o}_t &= \sigma\left( \mathbf{W}_o \mathbf{x}_{w_t} + \mathbf{U}_o \mathbf{h}_{t-1}  + \mathbf{b}_o  \right)\\
\mathbf{c}_t &= \mathbf{f}_t \odot  \mathbf{c}_{t-1}  +  \mathbf{i}_t  \odot \mathbf{\tilde{c}}_t\\
\mathbf{h}_t &= \mathbf{o}_t \odot \tanh \left( \mathbf{c}_t   \right),
\end{aligned}
\end{equation}
where $\mathbf{W}_s, \mathbf{U}_s \in \mathbb{R}^{d \times d}$ and $\mathbf{b}_s \in \mathbb{R}^{d}$ for $s \in \{f, i, \tilde{c}, o\}$ are parameters of LSTM cells. $\sigma (\cdot)$ is an element-wise sigmoid function, $\tanh (\cdot)$ is an element-wise hyperbolic tangent function, and $\odot$ is an element-wise multiplication.

The hidden state $\mathbf{h}_t $ is affine-transformed followed by a softmax function:
\begin{equation} \label{softmax}
\text{Pr} \left( w_{t+1} = k| w_{<t+1}  \right) = \frac{\exp{\left( \mathbf{v}_k^{\top}  \mathbf{h}_t + b_k\right) } }{\sum_{k'}^{} \exp{\left( \mathbf{v}_{k'}^{\top}  \mathbf{h}_t + b_{k'}\right) }},
\end{equation}
where $\mathbf{v}_k$ is the $k$-th column of a parameter matrix $\mathbf{V} \in \mathbb{R}^{d \times |V|}$ and $b_k$ is the $k$-th element of a bias vector $\mathbf{b} \in \mathbb{R}^{d}$. In the training phase, we minimizes the negative log-likelihood with stochastic gradient descent.

\section{Experimental Settings}

We test five different model architectures on the three English corpora. Each model has a unique word embedding method, but all models share the same LSTM language modeling architecture,  that has 2 LSTM layers with 200 hidden units, $d = 200$. Except for the character only model, weights in the language modeling part are initialized with uniform random variables between -0.1 and 0.1. Weights of a bidirectional LSTM in the word embedding part are initialized with Xavier initialization \cite{DBLP:journals/jmlr/GlorotB10}. All biases are initialized to zero. 

Stochastic gradient decent (SGD) with mini-batch size of 32 is used to train the models. In the first $k$ epochs, the learning rate is 1. After the $k$-th epoch, the learning rate is divided by $l$ each epoch. $k$ manages learning rate decay schedule, and $l$ controls speed of decay. $k$ and $l$ are tuned for each model based on the validation dataset.

As the standard metric for language modeling, perplexity ($\text{PPL}$) is used to evaluate the model performance. Perplexity over the test set is computed as
$
 \text{PPL} = \exp \left(- \frac{1}{N} \sum_{i=1}^{N} \log p_{ \left( w_i | w_{<i} \right)}  \right),
$
where $N$ is the number of words in the test set, and $p_{ \left( w_i | w_{<i} \right)}$ is the conditional probability of a word $w_i$ given all the preceding words in a sentence. We use Theano \shortcite{2016arXiv160502688short} to implement all the models.  The code for the models is available from \url{https://github.com/nyu-dl/gated\_word\_char\_rlm}.

\subsection{Model Variations}

\noindent \textbf{Word Only (baseline)}$\:\:$ This is a traditional word-level language model and is a baseline model for our experiments.

\vspace{4pt}
\noindent \textbf{Character Only}$\:\:$ 
This is a language model where each input word is represented as a character sequence similar to the C2W model in \cite{wang:2015}. The bidirectional LSTMs have 200 hidden units, and their weights are initialized with Xavier initialization. In addition, the weights of the forget, input, and output gates are scaled by a factor of $4$.   The weights in the LSTM language model are also initialized with Xavier initialization. All biases are initialized to zero. A learning rate is fixed at $0.2$.
	
\vspace{4pt}
\noindent \textbf{Word \& Character}$\:\:$ This model simply concatenates the vector representations of a word constructed from the character input $\mathbf{x}^{\text{char}}_{w_t}$ and the word input $ \mathbf{x}^{\text{word}}_{w_t}$ to get the final representation of a word $\mathbf{x}_{w_t}$ , i.e.,
\begin{equation}
\mathbf{x}_{w_t} = \left[ 
\begin{array}{c}
\mathbf{x}_{w_t}^{\text{char}} ; \mathbf{x}_{w_t}^{\text{word}}\\
\end{array}
\right].
\end{equation}
Before being concatenated,  the dimensions of $\mathbf{x}^{\text{char}}_{w_t}$ and  $\mathbf{x}^{\text{word}}_{w_t}$ are reduced by half to keep the size of $\mathbf{x}_{w_t}$ comparably to other models. 

\vspace{4pt}
\noindent  \noindent \textbf{Gated Word \& Character, Fixed Value}$\:\:$ This model uses a globally constant gating value to combine vector representations of a word constructed from the character input $\mathbf{x}^{\text{char}}_{w_t}$ and the word input $ \mathbf{x}^{\text{word}}_{w_t}$ as 
\begin{equation} \label{fixed_gate}
\begin{aligned}
\mathbf{x}_{w_t} &= \left( 1 -  g_{\text{}} \right)  \mathbf{x}^{\text{word}}_{w_t} + g_{\text{}}   \mathbf{x}^{\text{char}}_{w_t},
\end{aligned}
\end{equation} 
where $g_{\text{}}$ is some number between $0$ and $1$. We choose $g_{\text{}} = \{0.25, 0.5, 0.75\}$. 

\vspace{4pt}
\noindent \textbf{Gated Word \& Character, Adaptive}$\:\:$ This model uses adaptive gating values to combine vector representations of a word constructed from the character input $\mathbf{x}^{\text{char}}_{w_t}$ and the word input $ \mathbf{x}^{\text{word}}_{w_t}$ as the Eq~\eqref{gate}.

\begin{table}[t]
	\small
	\begin{tabular}{l l | c  c  c }
		\hline &  & Train & Validation & Test \\  \hline
		\multirow{2}{*}{PTB}&\# Sentences & 42k & 3k & 4k \\
		&\# Word & 888k & 70k & 79k \\
		\hline 
		\multirow{2}{*}{BBC}&\# Sentences & 37k & 2k & 2k \\
		&\# Word & 890k & 49k & 53k \\
		\hline 
		\multirow{2}{*}{IMDB}&\# Sentences & 930k & 153k & 152k \\
		&\# Word & 21M & 3M & 3M \\
		\hline
	\end{tabular}
	\caption{\label{datasize} The size of each dataset.}
\end{table}

\subsection{Datasets}

\noindent \textbf{Penn Treebank}$\:\:$ We use the Penn Treebank Corpus \cite{DBLP:journals/coling/MarcusSM94} preprocessed by Mikolov et al. \shortcite{DBLP:conf/interspeech/MikolovKBCK10}. We use 10k most frequent words and 51 characters. In the training phase, we use only sentences with less than 50 words.

\vspace{4pt}
\noindent \textbf{BBC}$\:\:$ We use the BBC corpus prepared by Greene \& Cunningham \shortcite{DBLP:conf/icml/GreeneC06}. We use 10k most frequent words and 62 characters. In the training phase, we use sentences with less than 50 words.

\vspace{4pt}
\noindent \textbf{IMDB Movie Reviews}$\:\:$  We use the IMDB Move Review Corpus prepared by Maas et al. \shortcite{DBLP:conf/acl/MaasDPHNP11}. We use 30k most frequent words and 74 characters.  In the training phase, we use sentences with less than 50 words. In the validation and test phases, we use sentences with less than 500 characters.

\subsection{Pre-training}
For the word--character hybrid models, we applied a pre-training procedure to encourage the model to use both representations. The entire model is trained only using the word-level input for the first $m$ epochs and only using the character-level input in the next $m$  epochs. In the first $m$ epochs, a learning rate is fixed at $1$, and a smaller learning rate $0.1$ is used in the next $m$ epochs. After the $2m$-th epoch, both the character-level and the word-level inputs are used. We use $m = 2$ for PTB and BBC, $m = 1$ for IMDB.

Lample et al. \shortcite{DBLP:journals/corr/LampleBSKD16} report that a pre-trained word lookup table improves performance of their word \& character hybrid model on named entity recognition (NER). In their method,  word embeddings are first trained using skip-n-gram \cite{DBLP:conf/emnlp/LingTAFDBTL15}, and then the word embeddings are fine-tuned in the main training phase.

\begin{center}
	\begin{figure*}[]
		\centering
		\subfigure[Gated word \& character.]{\includegraphics[width=0.42\linewidth]{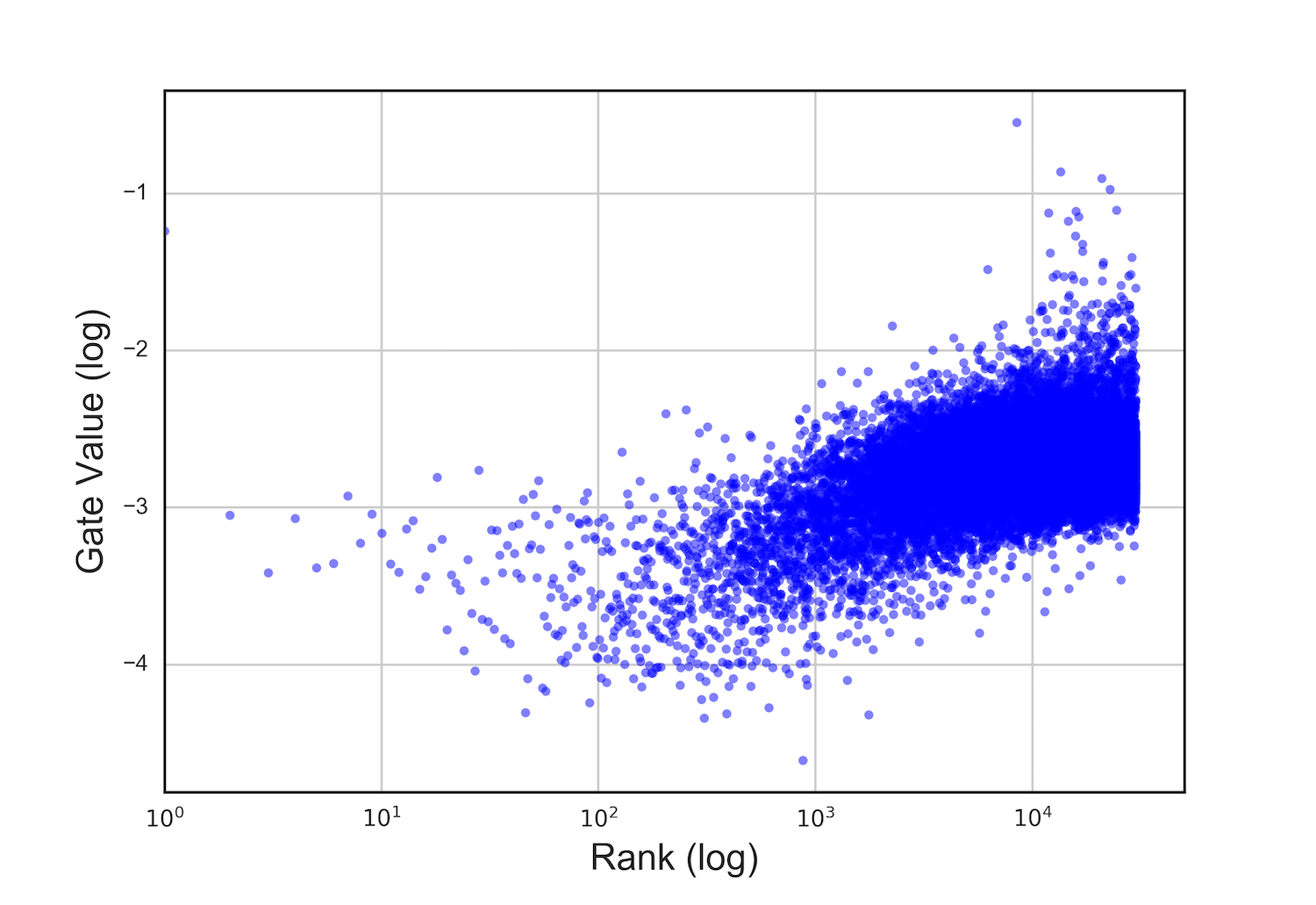}}\quad
		\hspace{24pt}
		\subfigure[Gated word \& character with pre-taining.]{\includegraphics[width=0.42\linewidth]{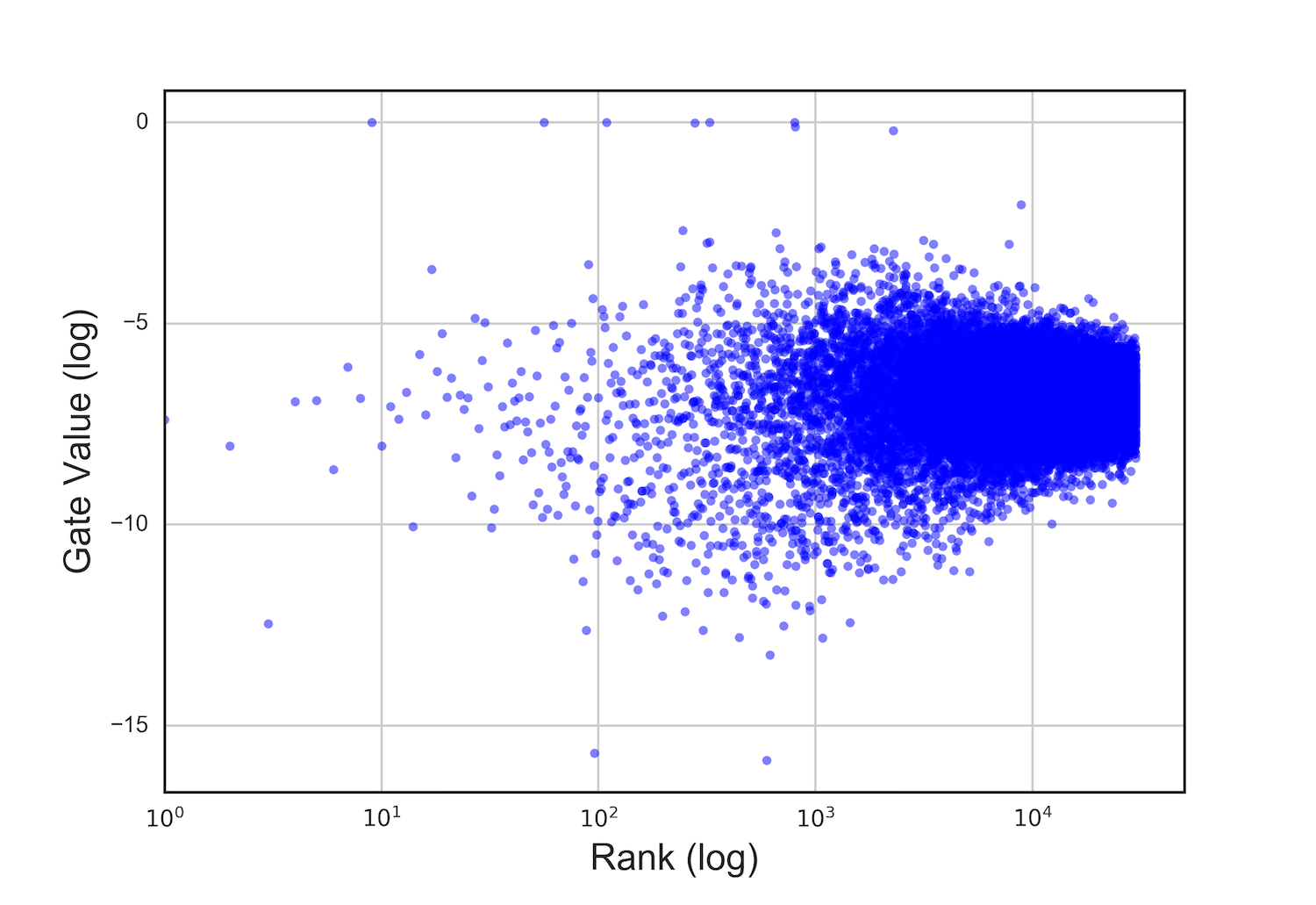}}
		\caption{A log-log plot of frequency ranks and gating values trained in the gated word \& character models with/without pre-training.}
		\label{fig:rank}
	\end{figure*}
\end{center}

\section{Results and Discussion}
\subsection{Perplexity}

Table~\ref{tab:perp} compares the models on each dataset. On the PTB and IMDB Movie Review dataset, the gated word \& character model with a fixed gating value, $g_{\text{const}}=0.25$, and pre-training achieves the lowest perplexity . On the BBC datasets, the gated word \& character model without pre-training achieves the lowest perplexity.

Even though the model with fixed gating value performs well, choosing the gating value is not clear and might depend on characteristics of datasets such as size. The model with adaptive gating values does not require tuning it and achieves similar perplexity.

\subsection{Values of Word--Character Gate}

The BBC and IMDB datasets retain out-of-vocabulary (OOV) words while the OOV words have been replaced by \texttt{<unk>} in the Penn Treebank dataset. On the BBC and IMDB datasets, our model assigns a significantly high gating value on the unknown word token \texttt{UNK} compared to the other words.

We observe that pre-training results the different distributions of gating values. As can be seen in Fig.~\ref{fig:rank} (a), the gating value trained in the gated word \& character model without pre-training is in general higher for less frequent words, implying that the recurrent language model has learned to exploit the spelling of a word when its word vector could not have been estimated properly.  Fig.~\ref{fig:rank} (b) shows that the gating value trained in the gated word \& character model with pre-training is less correlated with the frequency ranks than the one without pre-training. The pre-training step initializes a word lookup table using the training corpus and includes its information into the initial values. We hypothesize that the recurrent language model tends to be word--input--oriented if the informativeness of word inputs and character inputs are not balanced especially in the early stage of training.    

Although the recurrent language model with or without pre-training derives different gating values, the results are still similar. We conjecture that the flexibility of modulating between word-level and character-level representations resulted in a better language model in multiple ways.

Overall, the gating values are small. However, this does not mean the model does not utilize the character-level inputs.  We observed that the word vectors constructed from the character-level inputs usually have a larger L2 norm than the word vectors constructed from the word-level inputs do. For instance, the mean values of L2 norm of the 1000 most frequent words in the IMDB training set are $52.77$ and  $6.27$ respectively. The small gate values compensate for this difference.

\section{Conclusion}

We introduced a recurrent neural network language model with LSTM units and a word--character gate. Our model was empirically found to utilize the character-level input especially when the model encounters rare words.  The experimental results suggest the gate can be efficiently trained so that the model can find a good balance between the  word-level and character-level inputs.

\section*{Acknowledgments}

This work is done as a part of the course DS-GA 1010-001 Independent Study in Data Science at the Center for Data Science, New York University. KC thanks the support by Facebook, Google (Google Faculty Award 2016) and NVidia (GPU Center of Excellence 2015-2016). YM thanks Kentaro Hanaki, Israel Malkin, and Tian Wang for their helpful feedback.  KC and YM thanks the anonymous reviewers for their insightful comments and suggestions.

\newpage

\bibliography{emnlp2016}
\bibliographystyle{emnlp2016}

\end{document}